\documentclass[lettersize,journal]{IEEEtran}
\usepackage{amsmath,amsfonts}
\usepackage{algorithmic}
\usepackage{algorithm}
\usepackage{array}
\usepackage[caption=false,font=normalsize,labelfont=sf,textfont=sf]{subfig}
\usepackage{textcomp}
\usepackage{stfloats}
\usepackage{url}
\usepackage{verbatim}
\usepackage{graphicx}
\usepackage{cite}
\usepackage{utfsym,colortbl}
\begin{document}

\title{Gradient Domain Diffusion Models for Image Synthesis}

\author{Yuanhao Gong\\College of Electronics and Information Engineering, Shenzhen University, China.~~gong.ai@qq.com}

\markboth{Journal of \LaTeX\ Class Files,~Vol.~14, No.~8, August~2021}%
{Shell \MakeLowercase{\textit{et al.}}: A Sample Article Using IEEEtran.cls for IEEE Journals}


\maketitle

\begin{abstract}
Diffusion models are getting popular in generative image and video synthesis. However, due to the diffusion process, they require a large number of steps to converge. To tackle this issue, in this paper, we propose to perform the diffusion process in the gradient domain, where the convergence becomes faster. There are two reasons. First, thanks to the Poisson equation, the gradient domain is mathematically equivalent to the original image domain. Therefore, each diffusion step in the image domain has a unique corresponding gradient domain representation. Second, the gradient domain is much sparser than the image domain. As a result, gradient domain diffusion models converge faster. Several numerical experiments confirm that the gradient domain diffusion models are more efficient than the original diffusion models. The proposed method can be applied in a wide range of applications such as image processing, computer vision and machine learning tasks.
\end{abstract}


\section{Introduction}
\IEEEPARstart{G}{enerative} models, such as ChatGPT and Claude, have become increasingly popular across various fields in recent years. These deep learning-based models have been used in natural language processing, image and video generation, and music creation. Despite their success, there is much to be explored in this exciting area of research and development. One area that is currently being explored is the use of generative models in the creation of virtual worlds. By leveraging the power of artificial intelligence, these models can create highly-realistic and immersive environments with stunning detail, allowing users to experience new worlds like never before.

Another area where generative models are showing promise is in healthcare. By analyzing large amounts of patient data, these models can help medical professionals identify patterns and make more accurate diagnoses. Furthermore, these models can also be used to simulate the effects of different treatments, allowing doctors to make more informed decisions when it comes to patient care.

The potential applications of these models in the future are vast and varied, ranging from automated content creation to personalized recommendations based on user preferences. As such, there is growing interest in exploring the capabilities of these models and pushing the boundaries of what is possible with this technology. 
\begin{figure}
	\centering
\includegraphics[width=\linewidth]{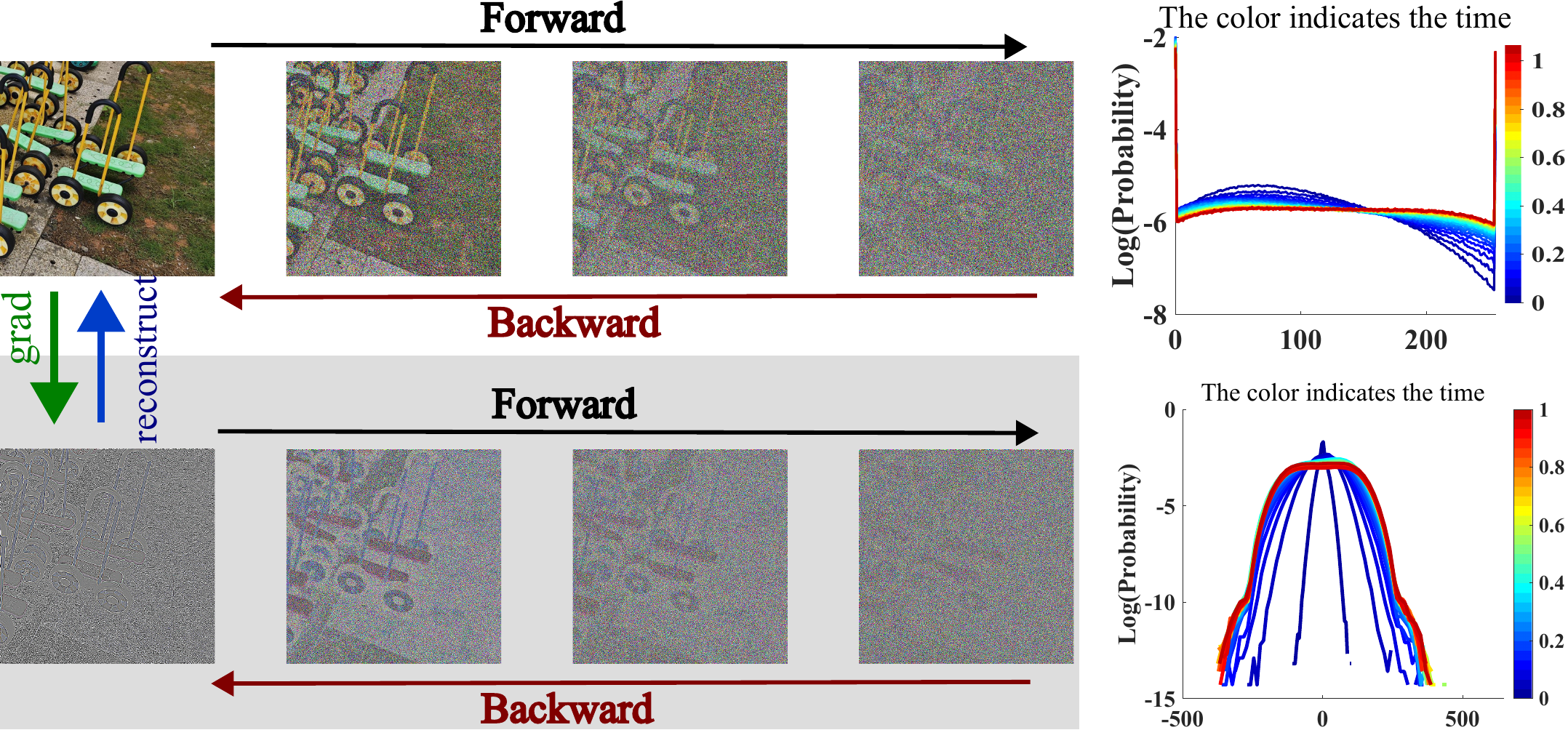}
	\caption{The top block is the diffusion model. The bottom shaded block is the gradient domain diffusion model. The green arrow indicates the gradient operator that brings the image into gradient domain. The blue arrow indicates the image reconstruction process from a gradient field (usually a Poisson solver neural network that transforms the gradient into an image). The right plots are their distribution behavior with time. Thanks to the sparsity of the gradient, the distribution in gradient domain converges faster.}
	\label{fig1}
\end{figure}
\subsection{Diffusion Model for Image Generation}
Diffusion models are a fascinating and versatile technology that is becoming increasingly popular for generating high-quality images and videos. These models use a process of iteratively diffusing information across a grid of pixels to generate an image or video. The result is a realistic and detailed representation of the original subject matter.

With the growing power of modern computing, diffusion models are becoming more accessible and widely used in a variety of fields, including film and video game production, scientific visualization, and even art. They have many advantages over other methods of generating images and videos.

Diffusion models are particularly useful in the creation of high-quality images and videos because they can capture fine details and subtle variations in color and light that might be missed by other methods. They can also generate images and videos that are more realistic and natural-looking than those produced by traditional methods.

One of the key advantages of diffusion models is that they are able to simulate the way light interacts with objects in the real world. This means that they can create images and videos that accurately reflect the way that light behaves in different environments, such as outdoors in natural light, or indoors under artificial lighting. This is crucial for creating realistic and accurate visual representations of objects and scenes.

Another advantage of diffusion models is that they are highly customizable. They can be tuned to produce images and videos with specific characteristics, such as a particular level of detail, a particular color palette, or a particular style. This makes them useful for a wide range of applications, from creating realistic renderings of architectural designs to generating abstract art.

Furthermore, diffusion models are a type of algorithm that generates high-quality images and videos by iteratively diffusing information across a grid of pixels. This method uses mathematical principles to simulate the diffusion of light and other physical phenomena, which makes it uniquely powerful and versatile.

As computing power continues to improve, diffusion models will become even more accessible and widely used. In the future, we can expect to see diffusion models applied to an even greater range of applications, further enhancing the quality and realism of images and videos.

\subsection{Gradient Domain Image Processing}
Gradient domain image processing is a fascinating technique used in image processing that operates on the gradients of an image as opposed to the image itself. These gradients are then used to manipulate the image in various ways, such as image smoothing, edge detection, and tone mapping. The beauty of this approach is that it allows for more precise and targeted editing of an image, giving the user greater control over the final product.

Moreover, gradient domain image processing has become increasingly popular in recent years due to its many applications. In addition to gradient domain compositing and inpainting, which were previously mentioned, this technique can also be used for image stitching. This involves combining multiple images into a single, larger image, which is particularly useful in fields such as landscape photography.

Another application of gradient domain image processing is image denoising, which involves removing noise from an image. This is accomplished by analyzing the gradients of the image and removing those that do not fit the pattern of the surrounding areas. This approach can be particularly useful in scientific imaging.

In addition to these applications, gradient domain image processing can also be used for color correction, image enhancement, and texture synthesis, among other things. With its many applications and benefits, it has become an essential tool in the field of image processing.

\subsection{Our contributions}
Due to the diffusion process, the diffusion models require a large number of time steps to generate images. To reduce the time steps, many distilling techniques are developed. However, such distilling methods introduce other issues, such as additional architecture design, retrain the network and even more failure cases. 

In this paper, we propose to perform the diffusion process in the gradient domain, which is much sparser than the intensity domain. Such sparsity leads to faster convergence. Our contributions include the following
\begin{itemize}
	\item  we present a novel gradient domain diffusion model.
	\item we show that gradient domain diffusion models converge faster than the image domain diffusion model.
\end{itemize}
\section{Gradient Domain Diffusion Models}
The forward diffusion process can be described by the stochastic differentiable equation
\begin{equation}
	\mathrm{d}x=f(x,t)\mathrm{d}t+h(t)\mathrm{d}w\,,
\end{equation} where $x$ is a sample, $f(x,t)$ is the drift coefficient, $h(t)$ is the diffusion coefficient and $w$ is a Wiener process.

Its reverse process can be described by 
\begin{equation}
	\mathrm{d}x=[f(x,t)-h^2(t)\nabla_x\log p_t(x)]\mathrm{d}t+h(t)\mathrm{d}w\,,
\end{equation} where $p_t(x)$ is the marginal probability density at time $t$. The score function $\nabla_x\log p_t(x)$ can be modeled by a parametric function $s_{\theta}(x,t)$ whose parameters can be optimized via score matching methods~\cite{Song2019}.
\subsection{Denoise Diffusion Probability Models}
The Denoise Diffusion probability model (DDPM) can be recovered with choosing the specific functions $f=-\frac{1}{2}\beta(t)x$ and $h=(\beta(t))^{-\frac{1}{2}}$. The forward step is
\begin{equation}
	x_t=\sqrt{1-\beta_t}x_{t-1}+\sqrt{\beta_t}\epsilon_{t-1}\,,
\end{equation} where $\epsilon_{t-1}$ satisfies a normal distribution $\epsilon_{t}\sim{\cal{N}}(0,\sigma^2)$. Thanks to the normal distribution and the first order Markov chain assumption, this process can have a closed form from the initial $x_0$ via
\begin{equation}
	x_t=\sqrt{\gamma_t}x_0+\sqrt{1-\gamma_t}\epsilon_0\,,
\end{equation}where $\gamma_t=\prod_{j=1}^{t}\alpha_j$ and $\alpha_t=1-\beta_t$.

Its inverse step is
\begin{equation}
	\begin{split}	
	x_{t-1}=&\frac{1}{\sqrt{\alpha_t}}[x_t-\frac{\beta_t}{\sqrt{1-\gamma_t}}\tilde{\epsilon}_{\theta}(x_t,t)]+\sqrt{\beta_t}\epsilon_{t}\\
	=&\frac{1}{\sqrt{\alpha_t}}[x_t-\frac{\beta_t}{\sqrt{1-\gamma_t}}\epsilon_{\theta}(x_t,t)]\,,
	\end{split}
\end{equation} where $\epsilon_{\theta}$ is a function with learn-able parameter $\theta$. Then, a neural network is trained to approximate the $\epsilon_{\theta}(x_t,t)$. 

The loss function is defined as
\begin{equation}
	{\cal L}(\theta)=\iiint\limits_{x_0,t,\epsilon_0}||\epsilon_0 -\epsilon_{\theta}(\sqrt{\gamma_t}x_0+\sqrt{1-\gamma_t}\epsilon_0,t)||^2\,.
\end{equation}
Since the DDPM is a special case of diffusion models, they have a close connection. The noise prediction and the score function have the relationship
\begin{equation}
	\label{eq:relationship}
\frac{\epsilon_{\theta}(x_t,t)}{\sqrt{1-\gamma_t}}\approx\nabla_x\log p_t(x)\,.
\end{equation}

\subsection{Gradient Domain}
We have used the notion $x$ to represent an image. Therefore, we use $\nabla x$ to denote the gradient of an image. For convenience, we omit the subscript for spatial coordinates if there is no ambiguity. For the given gradient field $\nabla x$, we try to find an image $\tilde{x}$ such that the following energy is minimized
\begin{equation}
	\label{eq:energy}
	{\cal E}(\tilde{x})=\frac{1}{2}\|\nabla \tilde{x}-\nabla x\|_2^2\,.
\end{equation} where the $\ell_2$ normal indicates the Gaussian measurement error.

\subsubsection{Convexity}
First of all, the above energy functional is convex. Its first derivative is 
\begin{equation}
	\frac{\partial{\cal E}}{\partial \tilde{x}}=\nabla^T(\nabla\tilde{x}-\nabla x)\,,
\end{equation} and its second derivative is
\begin{equation}
	\frac{\partial^2{\cal E}}{\partial \tilde{x}^2}=\nabla^T\nabla\,,
\end{equation} which is the standard Laplacian operator that is semi-positive definite. In such scenario, the Eq.~\eqref{eq:energy} is convex. Therefore, optimizing Eq.~\eqref{eq:energy} can use convex optimization methods.

\subsubsection{Uniqueness}
Minimizing Eq.~\eqref{eq:energy} is equivalent with solving a Poisson equation (with Dirichlet or Neumann boundary condition). More specifically, let $y=\nabla^T\nabla x$, then optimizing Eq.~\eqref{eq:energy} is to solve
\begin{equation}
	\label{eq:poisson}
	\left\{
	\begin{array}{ll}
		\Delta\tilde{x}=y\\
		\frac{\partial \tilde {x}}{\partial \vec{n}}|_{\partial \Omega}=0
	\end{array}\right.,
\end{equation} where $\Delta$ is the Laplacian operator, $\vec{n}$ is the normal direction and $\partial \Omega$ denotes the domain boundary. 

The above Poisson equation with Neumann boundary condition has an optimal unique solution. Let $\tilde{x}_1$ and $\tilde{x}_2$ be two solutions from above Poisson equation and $\phi=\tilde{x}_1-\tilde{x}_2$, then $\Delta \phi=0$ and $\frac{\partial \phi}{\partial \vec{n}}|_{\partial \Omega}=0$. According to Green’s First Identities, we have
\begin{equation}
	\int_{\Omega}((\nabla \phi)^2+\phi\Delta\phi)=\int_{\partial\Omega}\phi\frac{\partial \phi}{\partial\vec{n}}=0\,.
\end{equation} Therefore, $\nabla \phi=0$ and thus $\phi$ is a constant. The difference between $\tilde{x}_1$ and $\tilde{x}_2$ is constant, indicating the uniqueness of the equation. Such mathematical property highlights the theoretical advantage of gradient domain image processing.

\subsubsection{Sparsity}
\begin{figure}
\centering
\subfloat[image $x$]{\includegraphics[width=0.4\linewidth]{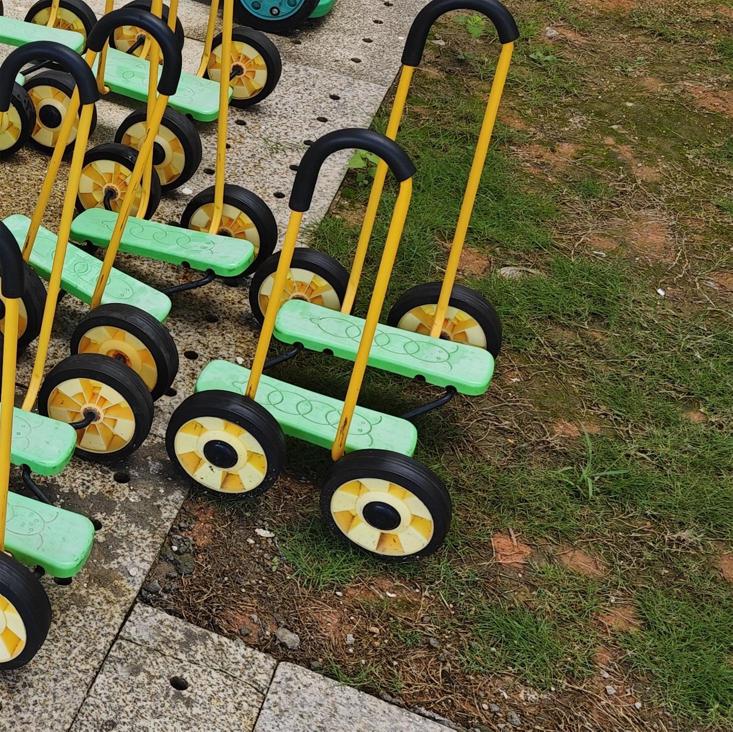}%
	\label{fig_first_case}}
\hfil
\subfloat[Laplacian $\Delta x$]{\includegraphics[width=0.4\linewidth]{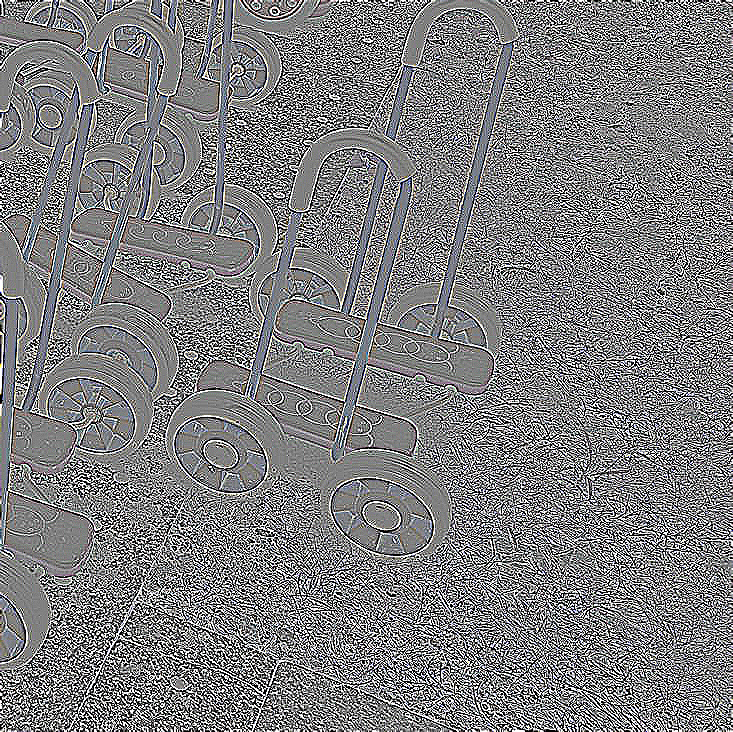}%
	\label{fig_second_case}}
	
	\subfloat[histogram of $x$]{\includegraphics[width=0.45\linewidth]{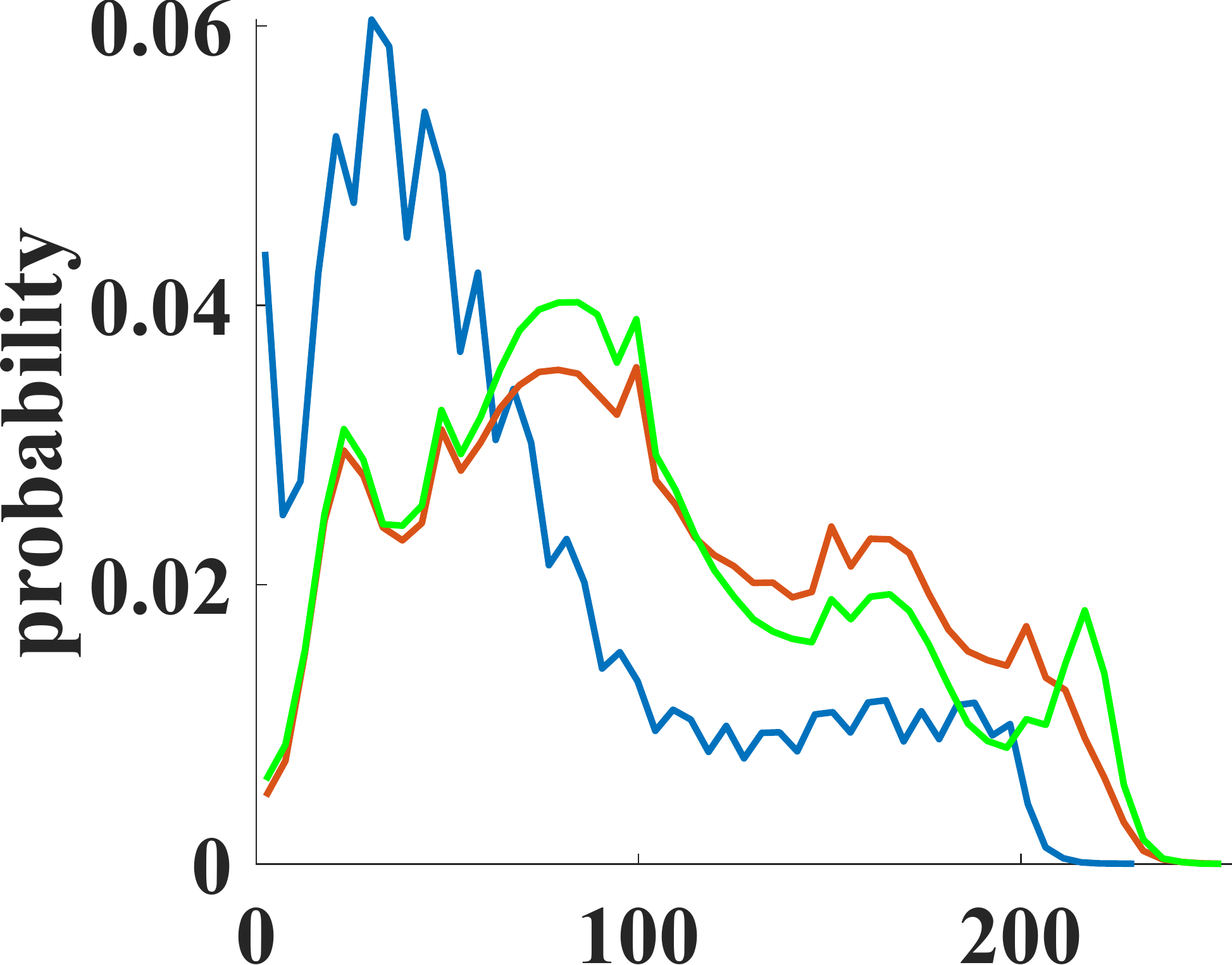}%
		\label{fig_firstH_case}}
	\hfil
	\subfloat[histogram of $\Delta x$]{\includegraphics[width=0.45\linewidth]{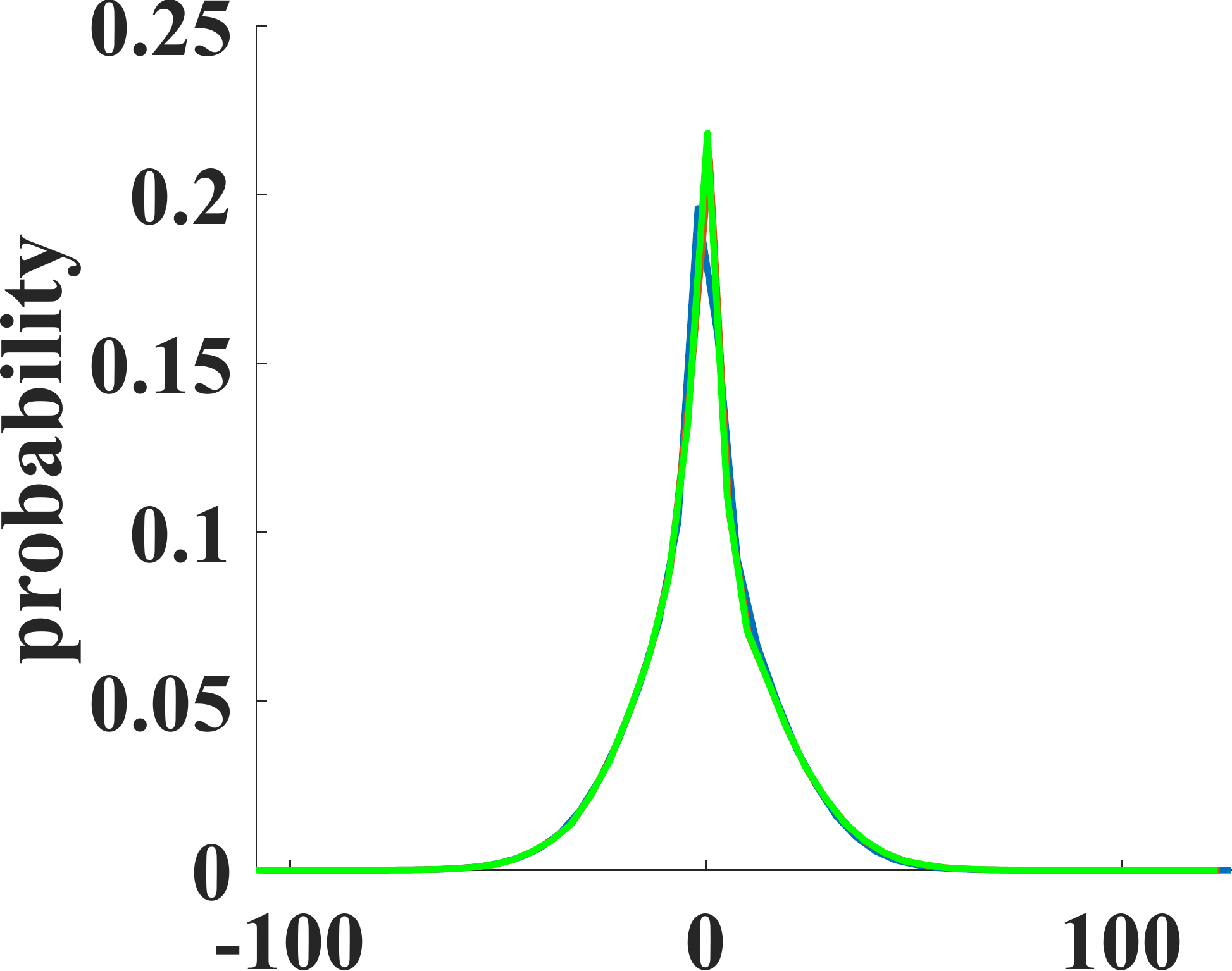}%
		\label{fig_secondH_case}}
		
		\subfloat[change with $t$]{\includegraphics[width=0.4\linewidth]{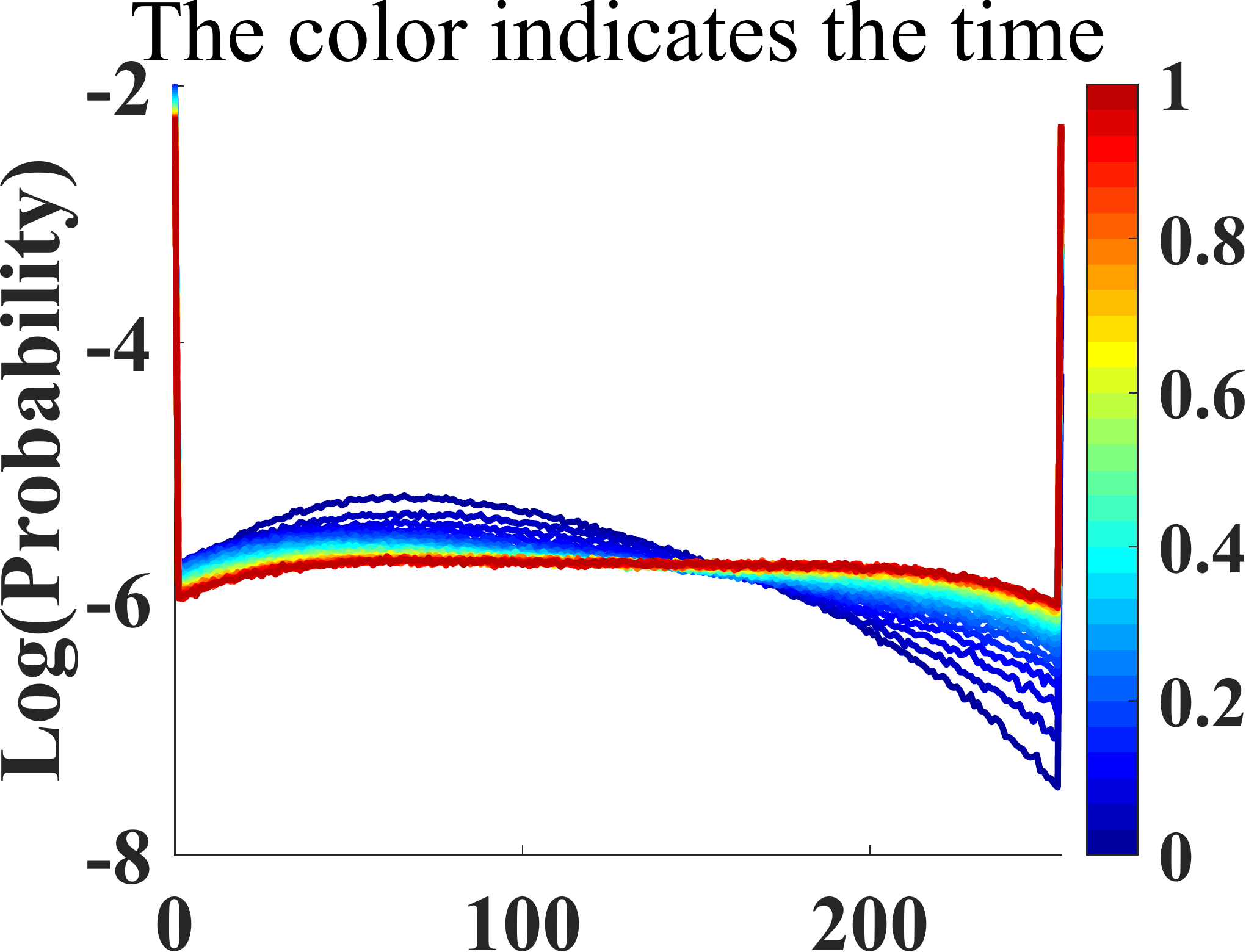}}%
		\hfil
		\subfloat[change with $t$]{\includegraphics[width=0.4\linewidth]{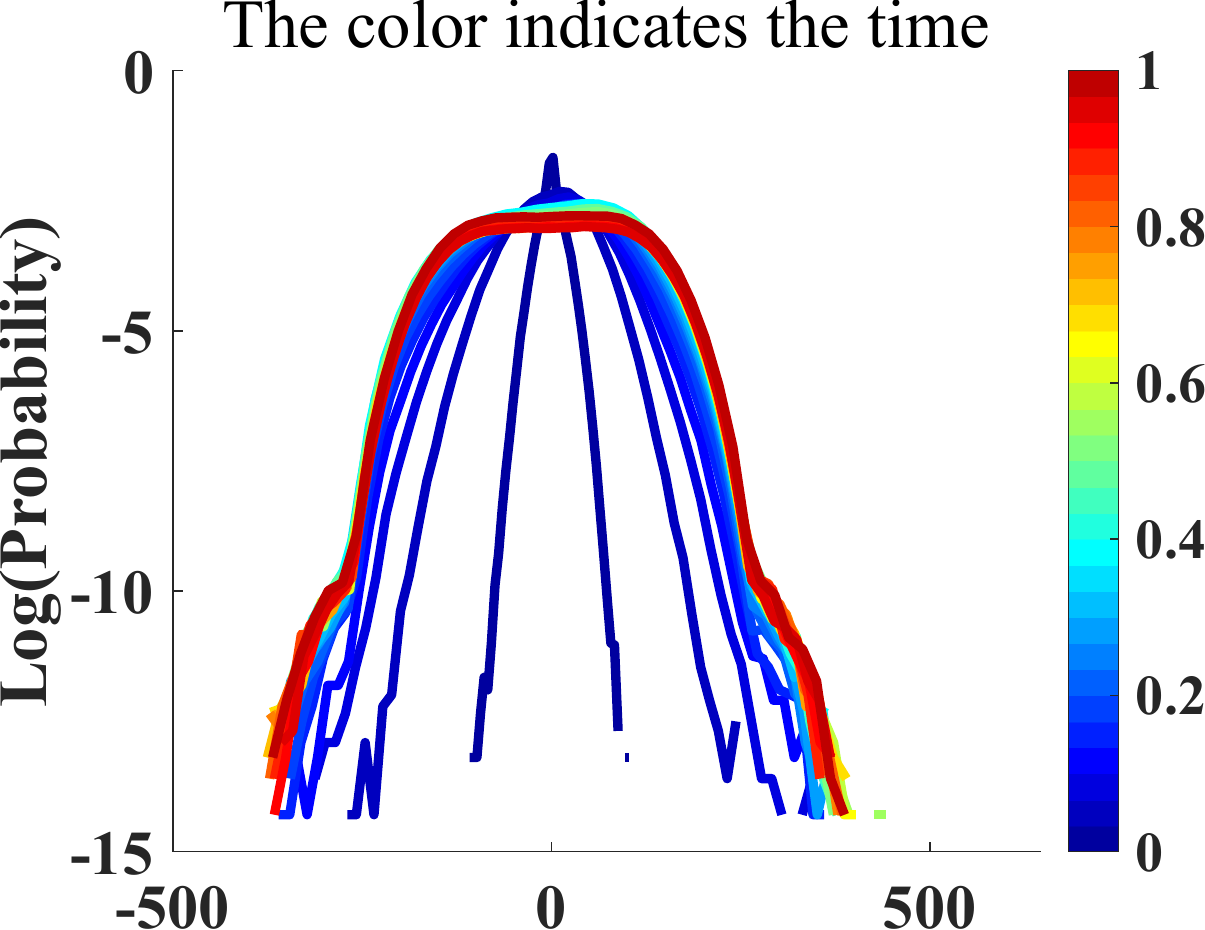}%
		}
\caption{(top) One image and its Laplacian field. (middle) their distributions. The color indicates the corresponding color channel. The Laplace is sparser. Moreover, different color channels satisfy almost the same distribution. (bottom) their changes in the log scale with the added noise, where the color indicates the time. The distribution from the Laplace field converges faster.}
\label{fig_sim}
\end{figure}

Besides the above uniqueness, the gradient domain is much sparser than the intensity domain because the gradient only captures the difference between neighbor intensities~\cite{gong:gdp}. In an image, the large gradient only appears around the edges. In contrast, zero gradient almost fills the whole imaging domain. Therefore, the probability $p(\nabla x=0)$ is quite high. One example is shown in Fig.~\ref{fig_sim}.

\subsection{Gradient Domain Diffusion Models}
Thanks to the sparsity in the gradient domain, the added noise becomes the dominant part. Therefore, the distribution converges faster to the normal distribution. More specifically, the forward step becomes
\begin{equation}
	\nabla x_t=\sqrt{\gamma_t}\nabla x_0+\sqrt{1-\gamma_t}\nabla\epsilon_0\,,
\end{equation}where $\gamma_t=\prod_{j=1}^{t}\alpha_j$ and $\alpha_t=1-\beta_t$. Since $\epsilon_{0}\sim{\cal{N}}(0,\sigma^2)$, the gradient also satisfies a normal distribution $\nabla_u\epsilon_{0}=\epsilon_{0}(u,v)-\epsilon_{0}(u-1,v)\sim{\cal{N}}(0,2\sigma^2)$. The gradient in the $v$ direction also has the same situation. Therefore, we have $\nabla\epsilon_{0}=\sqrt{2}\epsilon_{0}$ and above equation becomes
\begin{equation}
	\nabla x_t=\sqrt{\gamma_t}\nabla x_0+\sqrt{2(1-\gamma_t)}\epsilon_0\,.
\end{equation} In this equation, we directly add noise into the gradient field and the full forward process is in the gradient domain.

Moreover, the residual part in above equation is larger than the counterpart in the image domain $| \sqrt{2(1-\gamma_t)}\epsilon_0|\ge \sqrt{(1-\gamma_t)}\epsilon_0$. As a result, the neural network can capture the added noise easier in the gradient domain.

Its inverse step becomes
\begin{equation}
	\nabla x_{t-1}=\frac{1}{\sqrt{\alpha_t}}[\nabla x_t-\frac{\beta_t}{\sqrt{1-\gamma_t}}\nabla\epsilon_{\theta}(\nabla x_t,t)]\,.
\end{equation} Now, instead of using neural network to approximate $\epsilon_{\theta}(\nabla x_t,t)$ and then taking the gradient, we directly use a neural network $g_{\theta}(\nabla x_t,t)$ to approximate $\nabla\epsilon_{\theta}(\nabla x_t,t)$.
\begin{equation}
	g_{\theta}(\nabla x_t,t)=\nabla\epsilon_{\theta}(\nabla x_t,t)\,.
\end{equation}
Taking Eq.~\eqref{eq:relationship} into above, we have the following relationship
\begin{equation}
	g_{\theta}(\nabla x_t,t)\approx \sqrt{1-\gamma_t}\nabla[\nabla_x\log p_t(x)]\,.
\end{equation}
Therefore, the learned $g_{\theta}$ can be interpreted as the gradient of the score function with respect to the spatial coordinates. Therefore, the output of $g_{\theta}$ is also sparse.

We use the similar $\ell_2$ loss function in our model
\begin{equation}
	\label{eq:myloss}
	{\cal L}_g(\theta)=\iiint\limits_{x_0,t,\epsilon_0}|| \epsilon_{0}-g_{\theta}(\sqrt{\gamma_t}\nabla x_0+\sqrt{2(1-\gamma_t)}\epsilon_0,t)||^2\,.
\end{equation} The learned parameters $\theta$ are determined by the gradient $\nabla x$ instead of $x$. Both the forward and backward processes are in the gradient domain. Therefore, this model is named gradient domain diffusion model (GDDM).
\subsection{Laplace Domain Diffusion Models}
Noticing that the gradient domain eventually turns into the Laplace field in Eq.~\eqref{eq:poisson}, we can directly perform the diffusion process on the Laplace field. More specifically, the forward step becomes
\begin{equation}
	\Delta x_t=\sqrt{\gamma_t}\Delta x_0+\sqrt{1-\gamma_t}\Delta\epsilon_0\,,
\end{equation}where $\gamma_t=\prod_{j=1}^{t}\alpha_j$ and $\alpha_t=1-\beta_t$. 

The Laplacian from the noise also satisfies a normal distribution. For the discrete Laplacian operator, we use the following discrete finite element kernel to compute its Laplacian field 
\begin{equation}
	\label{eq:kernels}
	\small
	\begin{pmatrix} 0 & \frac{1}{4} & 0 \\ \frac{1}{4} & -1&\frac{1}{4} \\
		0 & \frac{1}{4} & 0
	\end{pmatrix}\,.	
\end{equation}
Since $\epsilon_{0}\sim{\cal{N}}(0,\sigma^2)$, its Laplacian field $\Delta\epsilon_{0}\sim{\cal{N}}(0,((\frac{1}{4})^2*4+1^2)\sigma^2)={\cal{N}}(0,(\frac{\sqrt{5}}{2}\sigma)^2)$. Therefore, the Laplacian field can be written as $\Delta\epsilon_0=\frac{\sqrt{5}}{2}\epsilon_{0}$. As a result, we can directly add the noise in the Laplacian domain and the forward step becomes
\begin{equation}
	\Delta x_t=\sqrt{\gamma_t}\Delta x_0+\frac{\sqrt{5(1-\gamma_t)}}{2}\epsilon_0\,.
\end{equation}
And the full forward process is in the Laplacian domain.

The noise part in this equaiton is larger than the counterpart in the gradient domain and intensity domain $|\frac{\sqrt{5(1-\gamma_t)}}{2}\epsilon_0|\ge|\sqrt{2(1-\gamma_t)}\epsilon_0|\ge |\sqrt{1-\gamma_t}\epsilon_0|$. This is another reason that the noise in the Laplacian domain quickly becomes dominant part.

Its inverse step becomes
\begin{equation}
	\Delta x_{t-1}=\frac{1}{\sqrt{\alpha_t}}[\Delta x_t-\frac{\beta_t}{\sqrt{1-\gamma_t}}\Delta\epsilon_{\theta}(\Delta x_t,t)]\,.
\end{equation} Now, we use a neural network to approximate $\Delta\epsilon_{\theta}(\Delta x_t,t)$.
\begin{equation}
	G_{\theta}(\Delta x_t,t)=\Delta\epsilon_{\theta}(\Delta x_t,t)\,.
\end{equation}
We use the similar $\ell_2$ loss function in our model
\begin{equation}
	\label{eq:myloss2}
	{\cal L}_G=\iiint\limits_{x_0,t,\epsilon_{0}}|| \epsilon_{0}-G_{\theta}(\sqrt{\gamma_t}\Delta x_0+\frac{\sqrt{5(1-\gamma_t)}}{2}\epsilon_0,t)||^2\,.
\end{equation} Since this model works in the Laplacian domain, we name this model as Laplacian domain diffusion model (LDDM).

\subsection{Why converge faster?}
As mentioned, the gradient and Laplacian fields are much sparser than the intensity domain. When adding the random noise, the noise quickly becomes the dominant part. Therefore, the gradient and Laplacian fields require much less steps to converge to a steady state than the diffusion process in the image intensity domain. Such behavior can be clearly seen in Fig.~\ref{fig_sim} (e) and (f).

Thanks to the sparsity, not only the training process converges faster, but also the sampling process becomes faster. In the backward (sampling) process, only the gradients at a small number of pixel locations are need to be recovered in the gradient domain. And the most of gradients in the output are zeros. 

\subsection{Poisson Network}
After generating an image gradient $\nabla x$ or Laplacian field $\Delta x$, we still need to recover $x$. Instead of solving the Poisson equation in Eq.~\eqref{eq:poisson}, we use a U-net neural network to recover the image. And this module is named Poisson network.

The Poisson network module is more robust than solving the Poisson equation because the generated gradient field might not be integratable. Therefore, we can train this Poisson network individually via adding random noise in the gradient or Laplacian field. Then we fix it for the backward diffusion process. This network not only accepts $\nabla x_0$ as input but also $\nabla x_{0.2}$ because it is tolerant of the noise.

\section{Experiments}
In this section, we numerically show that the diffusion process in the gradient and Laplacian domain converges faster than the process in the image domain. Therefore, the training process is much easier than the traditional diffusion models. Moreover, the generation process is also much faster because the backward process only generates edges of the image.

The generation process is similar to painting. We first draw the sketch or lines of the whole image and then fill in each region with more details. The backward diffusion process is the sketching process, drawing the contours of the image. Then the Poisson network is to fill the region with more details.
\subsection{Converge Faster}
\begin{figure}
	\centering
	\subfloat[image domain]{\includegraphics[width=0.4\linewidth]{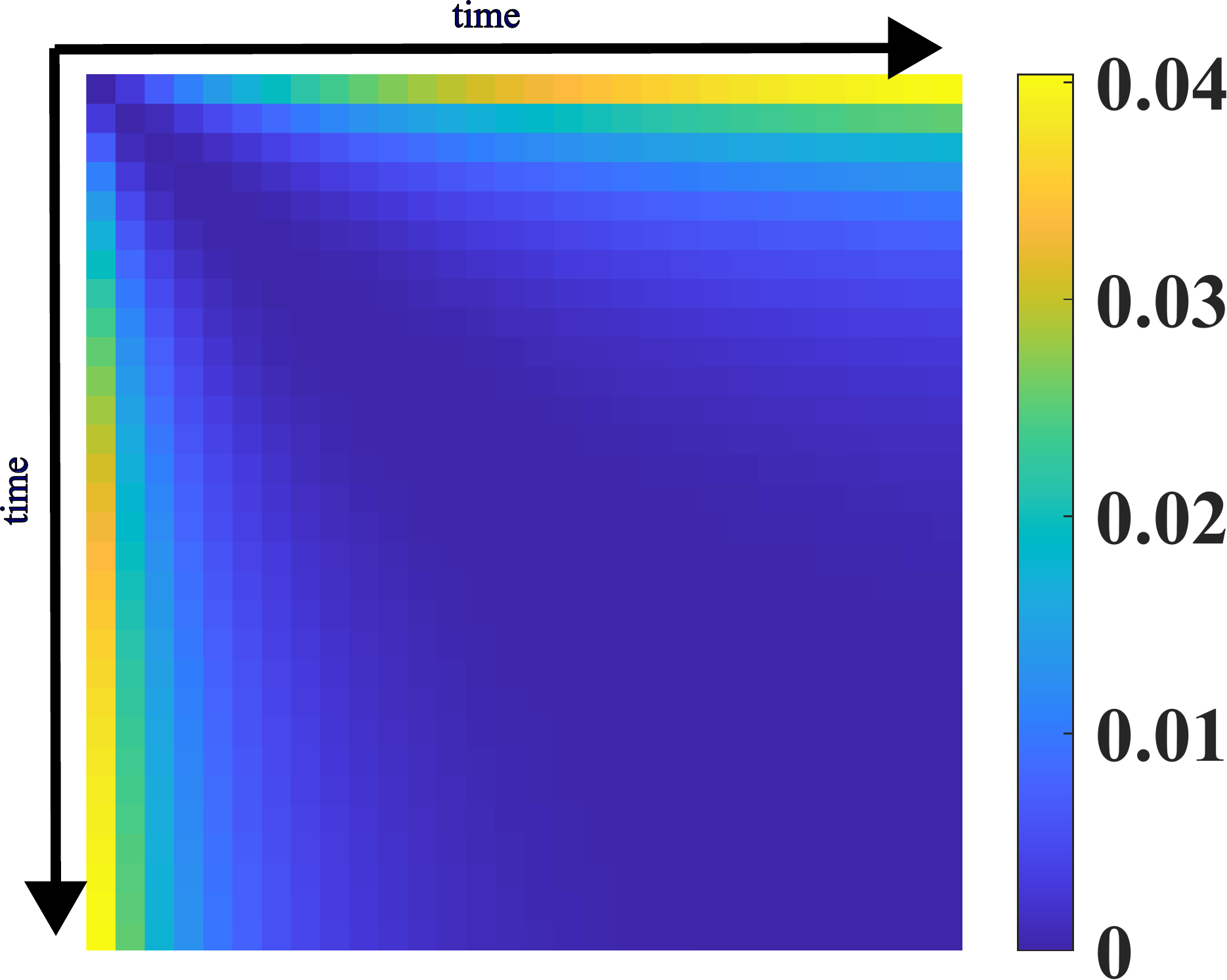}}
	\hfil
	\subfloat[Laplacian domain]{\includegraphics[width=0.4\linewidth]{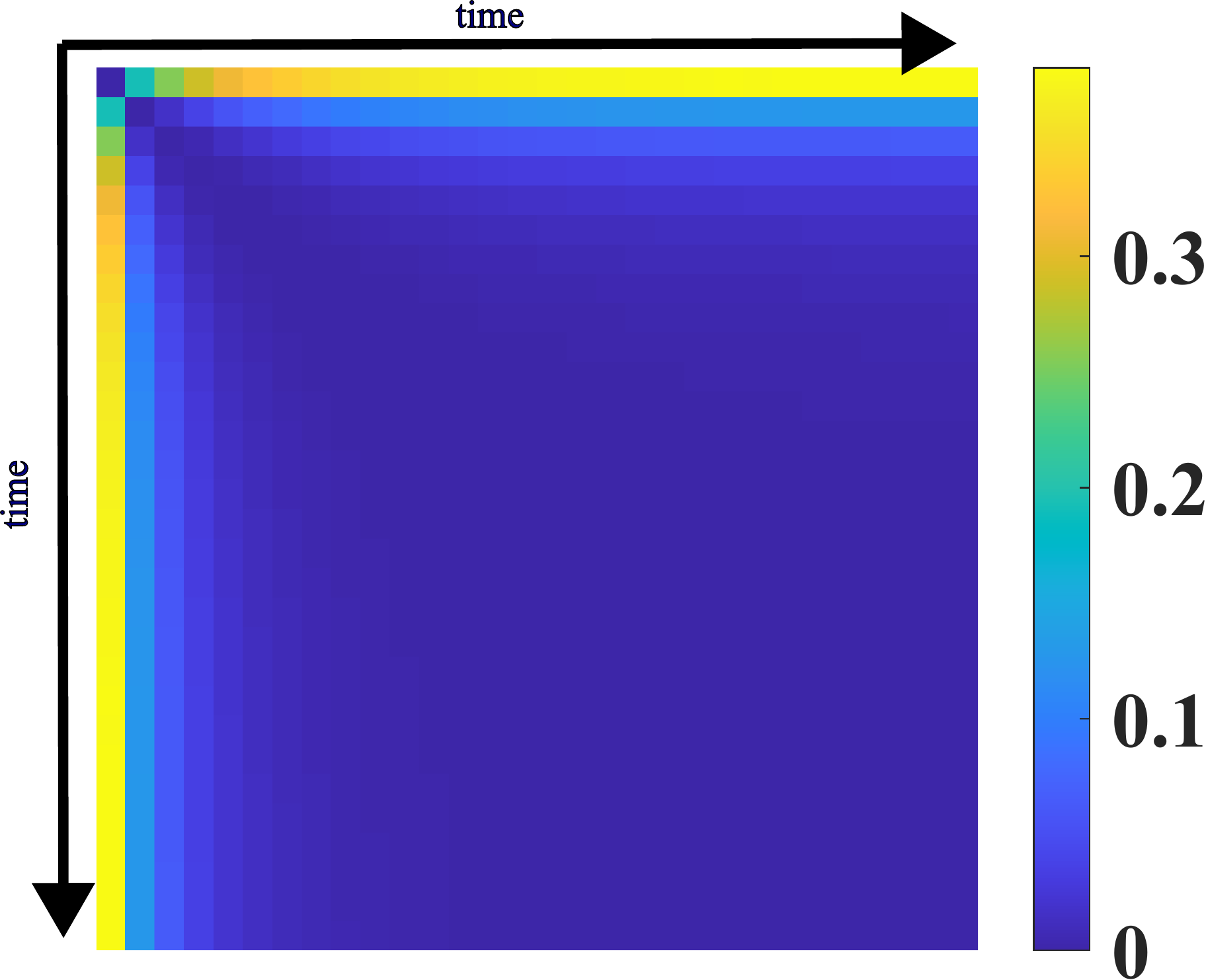}}
	
	\subfloat[first line in (a)]{\includegraphics[width=0.4\linewidth]{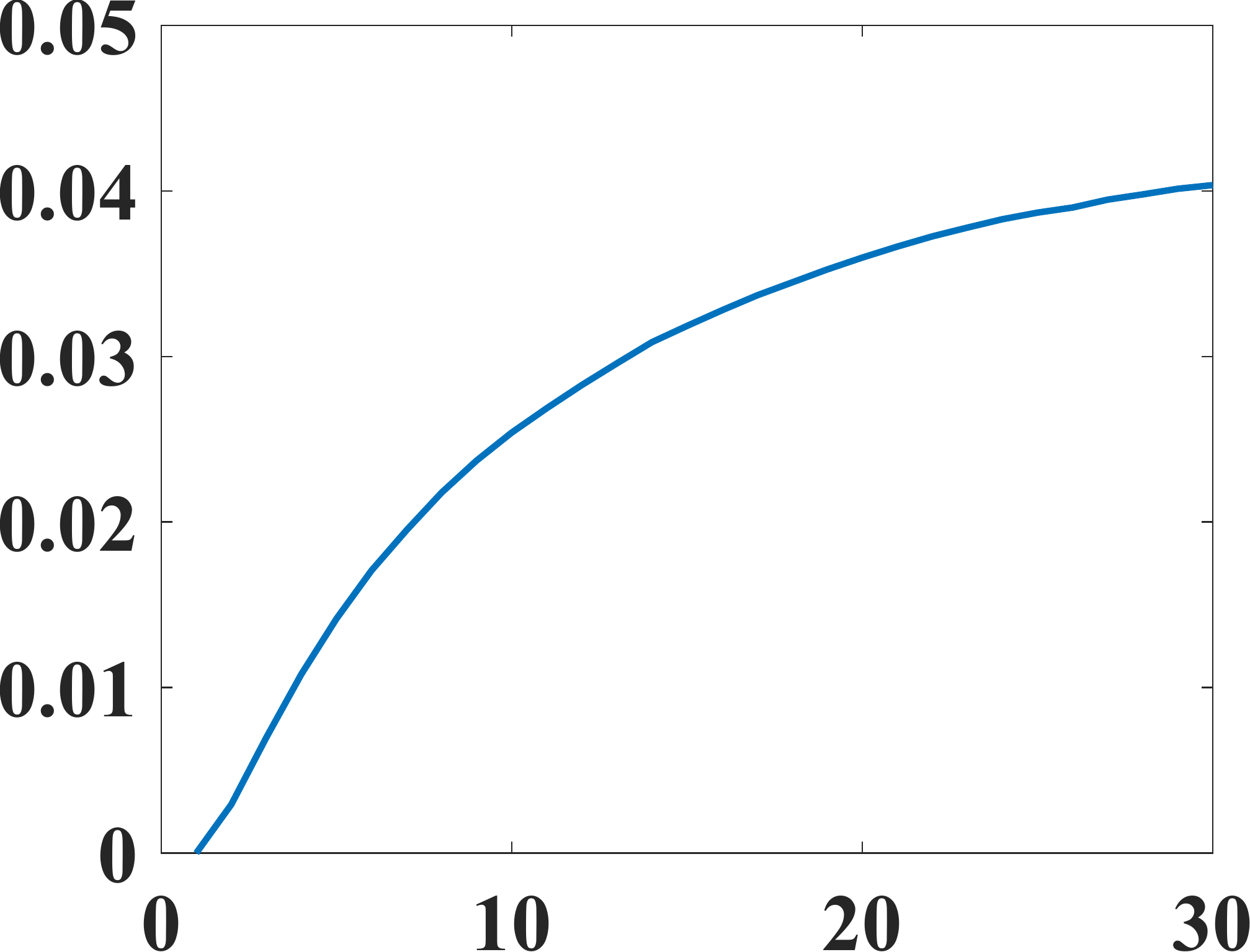}}
	\hfil
	\subfloat[first line in (b)]{\includegraphics[width=0.4\linewidth]{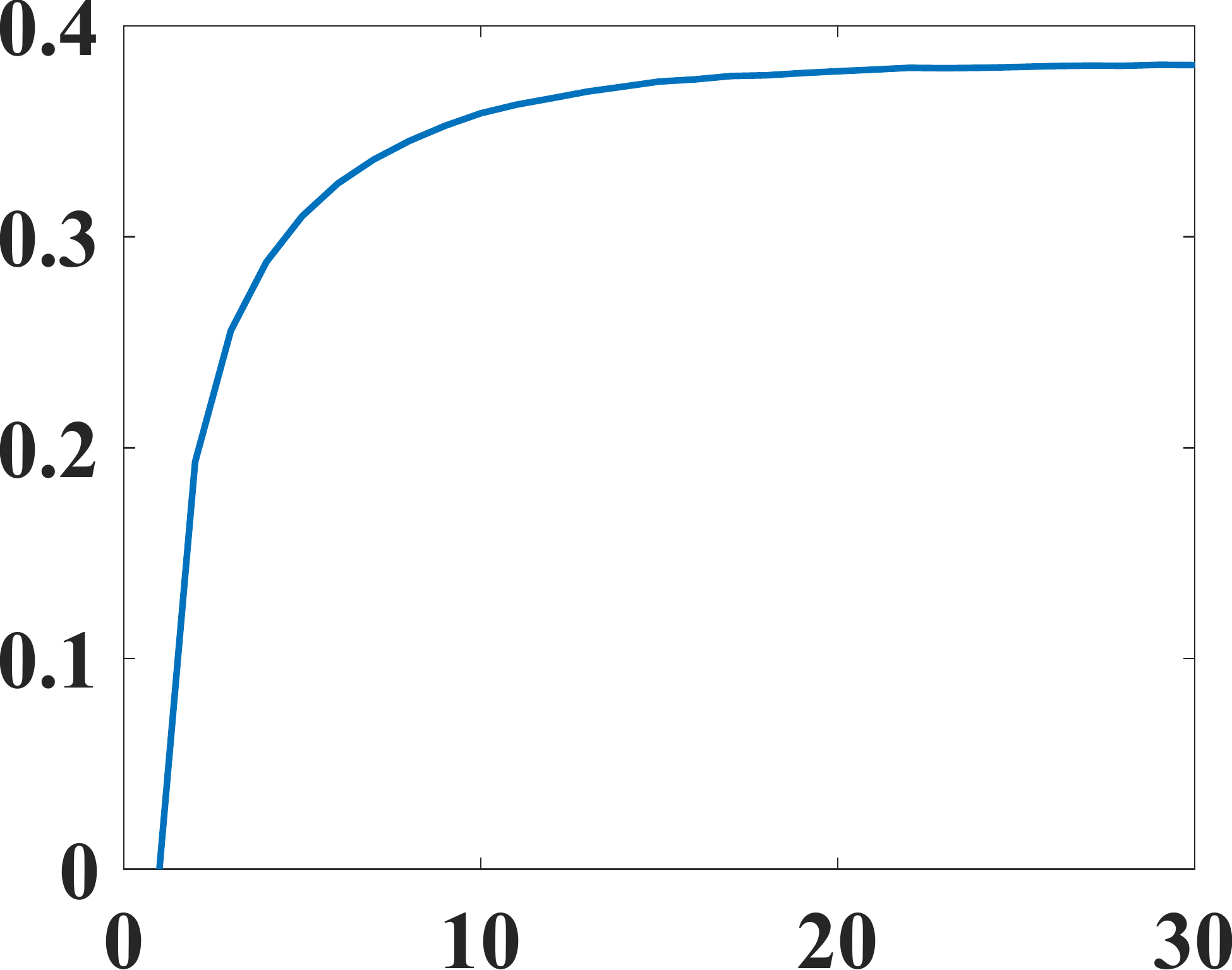}}
	\caption{(a) the Jensen-Shannon Divergence in the image domain. (b) Jensen-Shannon Divergence in the Laplacian domain. Bottom panels are their first line plot, showing the convergence speed.}
	\label{fig_sim2}
\end{figure}

To show that the gradient and Laplacian field converges faster, we compute the Jensen-Shannon Divergence between distributions at different time steps on the illustration image from Fig.~\ref{fig_sim} (a). The arrows in Fig.~\ref{fig_sim2} (a) and (b) indicate the forward time steps.

As shown in Fig.~\ref{fig_sim2}, the Jensen-Shannon Divergence in the Laplacian domain converges faster than in the image domain. Their first line plots in Fig.~\ref{fig_sim2} (c) and (d) clearly shows such difference. Thanks to the faster convergence, the training ans sampling process in the gradient and Laplacian domain requires much less time steps than the image intensity domain.
\subsection{Guided Sparsity}
The loss function Eq.~\eqref{eq:myloss} in GDDM and the loss function Eq.~\eqref{eq:myloss2} in LDDM do not force the sparsity for the output $\nabla\tilde{x}$ and $\Delta\tilde{x}$. To impose such sparsity, we can add a reconstruction error term to guide the backward process in the loss functions, leading to 
\begin{equation}
	{\cal L}_g+\lambda ||\nabla\tilde{x}-\nabla x||^2\,,
\end{equation}
and
\begin{equation}
	{\cal L}_G+\lambda ||\nabla\tilde{x}-\nabla x||^2\,,
\end{equation} respectively.

Other guidance such as spatial attention can also be imposed via the attention mechanism for better describing the categories such as human face, building, and flowers. 

\section{Conclusion}
In this paper, we have introduced a new method that performs diffusion processes in the gradient and Laplacian domains. Our approach involves using the well-known Poisson equation to reconstruct the image from the gradient and Laplacian fields. This not only guarantees the mathematical soundness of the diffusion process but also enhances the quality of image reconstruction.

To support our claims, we have demonstrated that the sparsity of the image in the gradient and Laplacian domains leads to faster convergence rates. We have presented the mathematical equations and numerical validation that demonstrate the effectiveness of our approach. 

Such sparsity leads to more efficient learning process (forward) and sampling process (backward). Therefore, the proposed method does not require a large number of time steps to generate an image.

Moreover, we have utilized a Poisson network to reconstruct the image from the gradient or Laplacian fields generated. This network ensures the robustness of the reconstruction process and enables the acceptance of noisy input.

The backward process can be considered to generate contours or sketch of an image. The Poisson network can be considered to fill the contours or sketch with more details. Such procedure is similar to the way that human artist paints on the paper.  

Our approach has the potential to achieve more accurate and efficient diffusion processes in the gradient and Laplacian domains. Furthermore, the use of the Poisson network can have widespread implications in other domains where diffusion processes are employed. Overall, our findings suggest that this approach can be a valuable addition to the existing techniques in this field.

In conclusion, the proposed gradient domain diffusion models have mathematical supports and numerically converge faster. They will play an important role in neural networks and machine learning, such as image inpainting, restoration, segmentation, synthesis, ect~\cite{chenouard:2014,gong2009symmetry,Lewis2019,zhao2023survey,Gong2012,Brown2020,gong2013a,Yu2019,Gong:2014a,Yin2019a,gong:phd,Yu2022a,gong:gdp,Guo2022,gong:cf,Zong2021,gong:Bernstein,Ezawa2023,Gong2017a,Tang2021a,Gong2018,Gong2018a,Yu2020,GONG2019329,Sancheti2022,Gong2019a,Tang2021,Gong2019,Yin2019b,Gong2022,Yin2020,Gong2020a,Jin2022,Gong2021,Tang2022,Gong2021a,Tang2022a,Tang2023,Gong2022,Tang2023a,Xu2023,Han2022,Gong2023a,Scheurer2023,Gong2023,Zhang2023b}.
\bibliographystyle{IEEEtran}
\bibliography{IEEEabrv,../../IP}

\vfill

\end{document}